\documentclass{article}

\usepackage[preprint]{neurips_2020}

\usepackage[utf8]{inputenc} 
\usepackage[T1]{fontenc}    
\usepackage{hyperref}       
\usepackage{url}            
\usepackage{booktabs}       
\usepackage{amsfonts}       
\usepackage{nicefrac}       
\usepackage{microtype}      
\usepackage{amsmath}
\usepackage{amssymb}
\usepackage{bm}
\usepackage{float}
\usepackage{graphicx}
\usepackage{caption}
\usepackage{lipsum}
\title{\textsc{HyperGrid}: \\ Efficient Multi-Task Transformers with Grid-wise Decomposable Hyper Projections}

\author{%
 Yi Tay, Zhe Zhao, Dara Bahri, Donald Metzler, Da-Cheng Juan\\ 
 Google Research \\
 Mountain View, California \\ 
  \texttt{\{yitay,zhezhao,dbahri,metzler,dacheng\}@google.com} \\
}

\begin{document}

\maketitle

\begin{abstract}
Achieving state-of-the-art performance on natural language understanding tasks typically relies on fine-tuning a fresh model for every task. Consequently, this approach leads to a higher overall parameter cost, along with higher technical maintenance for serving multiple models. Learning a single multi-task model that is able to do well for all the tasks has been a challenging and yet attractive proposition. In this paper, we propose \textsc{HyperGrid}, a new approach for highly effective multi-task learning. The proposed approach is based on a decomposable hypernetwork that learns grid-wise projections that help to specialize regions in weight matrices for different tasks. In order to construct the proposed hypernetwork, our method learns the interactions and composition between a global (task-agnostic) state and a local task-specific state. We apply our proposed \textsc{HyperGrid} on the current state-of-the-art T5 model, demonstrating strong performance across the GLUE and SuperGLUE benchmarks when using only a single multi-task model. Our method helps bridge the gap between fine-tuning and multi-task learning approaches.
\end{abstract}

\section{Introduction}
Learning a single multi-task model that performs well across multiple targeted tasks is an attractive proposition for many reasons \citep{kaiser2017one,ruder2017overview,clark2019bam}. Although extremely challenging, this paradigm enables a substantial savings in overall parameter costs, along with eliminating the need for maintaining multiple models in production \citep{stickland2019bert}. However, achieving state-of-the-art performance on natural language understanding benchmarks today \citep{wang2018glue,wang2019superglue} still relies on fine-tuning a new model for every single task. This methodology is infeasible in many situations. Moreover, certain tasks rely on an extensive ensemble of models and/or task-specific fine-tuning tricks \citep{liu2019roberta,devlin2018bert,clark2020electra}. 

The single-task fine-tuning paradigm is well-established to be the dominant approach \citep{raffel2019exploring}, as training multiple tasks using a single set of parameters can be problematic in many ways, such as catastrophic forgetting \citep{french2002using,mccloskey1989catastrophic,mcclelland1995there,kirkpatrick2017overcoming} or the inherent difficulty of finding a consistently good model for all tasks. Inevitable task conflicts and difficulty in fitting all models within a set of hard parameters is also a challenging problem for multi-task co-training.


In this paper, we propose \textit{Gridwise Decomposable Hyper Projections} (\textsc{HyperGrid}), a new adaptive hypernetwork-based \citep{ha2016hypernetworks} projection layer that aims to improve multi-task learning performance in natural language understanding. Our goal is to obtain competitive performance on multiple tasks with a single model. Our eventual goal is to dispense with task specific fine-tuning tricks altogether. While neural networks typically maintain the same consistent set of parameters for all input instances, the proposed \textsc{HyperGrid} introduces instance-specific parameters by conditioning on the current input. This setup enables our model to learn task-specific reparameterization for each input instance, which mitigates several challenges of multi-task co-training. 

Our proposed \textsc{HyperGrid} belongs to a family of hypernetworks \citep{ha2016hypernetworks}, in which a side network is responsible for weight generation for the main network. In our case, task-conditioned hypernetworks provide greater flexibility and expressiveness for capturing the dynamics of multiple tasks within a single set of parameters. Specifically, we introduce two novel algorithmic improvements over the existing methods. 

First, we introduce the notion of grid-wise projections in which we assume a structural layout in vanilla projection layers. For each input sample, our grid-wise projections dynamically control the parameters in a grid-wise, region-specific manner. The structural segmentation of feed-forward layers is similar in spirit to mixture-of-experts gating \citep{shazeer2017outrageously}, albeit at a lower-level. Conversely, standard hypernetworks only consider row-wise re-weighting of weight matrices.

Second, we introduce \textit{decomposable} hyper-projections. The key idea is to learn rich compositional and pairwise interactions between dual hypernetworks. A dual setup is adopted, where we explore different hypernetwork composition variants. We introduce a novel local-global setup, which composes a local instance-specific and task-specific hyper-projection with a task agonstic global state embedding. This is intuitive since this setup is not only highly expressive and flexible but also serve as a factorization of local and global components. To the best of our knowledge, our work is the first to explore this setup with respect to learning conditional parameters.

In our experiments, we equip state-of-the-art pretrained Transformer models \citep{vaswani2017attention} with our proposed \textsc{HyperGrid} layers during fine-tuning. Specifically, we imbue the state-of-the-art Text-to-Text Transformers (T5) \citep{raffel2019exploring} with \textsc{HyperGrid}. Although the T5 model is already setup to be a good candidate for multi-task learning with little effort, models are still fine-tuned on individual tasks separately during GLUE/SuperGLUE evaluation since they perform better in this setup. Therefore, our proposed \textsc{HyperGrid} projection layers were designed to bridge the gap between multi-task co-training and task-specific fine-tuning. 

On a whole, our final result (on the test set) is able to match the performance of individually fine-tuned T5 with only a single model that is learned to fit all GLUE and SuperGLUE tasks at once. Moreover, we also outperform strong competitors that employ aggressive ensembling and task-specific tricks \citep{liu2019roberta,clark2020electra} with only a single model on all 16 tasks.


\section{Related Work}
Multi-task learning (MTL) \citep{caruana1997multitask} is a long standing research problem. Learning a single unified model that does well on multiple tasks is an uphill battle given well-known problems such as catastrophic forgetting \citep{kirkpatrick2017overcoming}. As such, learning a large number of tasks with a single set of model parameters is an extremely challenging endeavour. Moreover, the disproportionate amount of data per task is also potentially problematic \citep{lee2017fully,pfeiffer2020mad}, which results in models overfitting on high resource tasks but underfitting on low resource tasks.

Early work in multi-task NLP typically considered a hierarchical taxonomy of tasks \citep{hashimoto2016joint} where a clear hierarchy of tasks exist, such as POS $\rightarrow$ Chunking $\rightarrow$ entailment. The Joint Many-Task (JMT) model explores an incremental and hierarchical paradigm for building multi-task NLP models. Similarly,  \citep{sanh2019hierarchical} proposed a hierarchical multi-task model based on the intuition of low-level and high-level tasks. Another line of recent work explores casting all tasks into a form of question answering problem \citep{mccann2018natural} and using an interpolated pointer-generator \citep{see2017get} mechanism for generating `answers'.

Exploiting task relatedness as a means for improved model quality has been frequently explored. In relatively recent work, \citep{liu2019multi} proposed MTDNN, a multi-task deep neural network that shares parameters between several NLP tasks. The model achieves strong performance on the GLUE benchmark. However, MTDNN simply leverages MTL as a form of pretraining and uses task-specific models for final evaluation. The recent T5 (Text-to-Text Transfer Transformers) model \citep{raffel2019exploring} frames all NLP problems as a Seq2Seq \citep{sutskever2014sequence} problem. However, the best results are again obtained by task-specific fine-tuning.

Orthogonal to other research efforts, \citep{clark2019bam} proposed Born Again Neural Networks (BAM), a clever way to obtain a single multi-task network by knowledge distillation. \citep{stickland2019bert} proposed Projected Attention Layers for task-specific fine-tuning of BERT \citep{devlin2018bert}. \citep{zaremoodi2018adaptive} proposed Adaptive Knowledge Sharing\footnote{The authors of \citep{raffel2019exploring} explored this approach but did not find it to be satisfactory.} for low-resource neural machine translation. Our work is related to the literature surrounding hypernetworks \citep{ha2016hypernetworks} which have been found to useful in areas such as continual learning \citep{von2019continual}. Learning task-adaptive parameters to avoid catastrophic forgetting has also been a go-to strategy for continual learning \citep{yoon2019oracle}. Outside of the NLP domain, flexible parameter sharing approaches are also dominant strategies for learning multi-task models \citep{ma2018modeling,ma2019snr}. 

The key novelty behind our work lies in the decomposable and factorized formulation in which we leverage the composition of two (local and global) hypernetworks. Additionally, the grid-wise gating of transform layers is also new. This sets it apart from previous soft parameter sharing \citep{ma2018modeling,ma2019snr} and hypernetwork \citep{von2019continual,ha2016hypernetworks} based approaches.

\section{The Proposed Method}
This section outlines the key idea of the proposed algorithm. 
\subsection{The HyperGrid Projection Method}
\textsc{HyperGrid} operates on weight matrices (linear transformations), i.e., $Y = \bm{W}X +b$. In a hypernetwork formulation, instead of letting $\bm{W}$ be free weights, we generate $\bm{W}$ using a parameterized side network $H(.)$. 
\begin{align}
Y = \bm{W}x +b  \:\:\:\:\text{where}\:\:\:\: \bm{W}=H(X)
\end{align}
where $\bm{W} \in \mathbb{R}^{d_{m} \times d_{f}}$. In the case where $X$ is a single vector $\in \mathbb{R}^{d_m}$, we may parameterize $H(.)$ with a simple feed-forward layer. 
\begin{align}
H(X) = \sigma(\bm{U}X)\bm{1}^{\top} \odot \bm{W}   
\end{align}
where $\bm{1}$ is a column vector of ones, $\sigma$ is the sigmoid activation function and $U \in \mathbb{R}^{d_m \times d_f}$. The key idea the hypernetwork generates a vector, i.e., $\bm{U}X \in \mathbb{R}^{d_f}$ that is broadcast (multiplied by $\bm{1}$) and multiplied by $\bm{W}$, acting as a row-wise scaling of $\bm{W}$. We are also able to reduce $U \in \mathbb{R}^{d_m \times n}$ where $d_f \mod n =0$ and repeat the vector $\frac{d_f}{n}$ times to form the original dimension of $d_f$. These methods only consider scaling one dimension of $W$ (e.g., row-wise). We now consider methods beyond simple row-wise weight scaling.
\subsubsection{Decomposable Gridwise Projections} 
\begin{minipage}{0.58\linewidth}
In our method, we propose grid-wise projections that segments $\bm{W}$ into a grid, i.e., blocks of $\frac{d_m}{d_r} \times \frac{d_f}{d_c}$. We generate blocks by the outer product of $L_r \in \mathbb{R}^{d_r}$ and $L_c \in \mathbb{R}^{d_c}$. Note that $d_r$ and $d_c$ are user-specific hyperparameters that control the grid-size for the fan-in and fan-out of the output matrix. For simplicity, we consider divisible blocks where $d_r<d_m, d_m \mod d_r=0$ and $d_c<d_f, d_f \mod d_c=0$. In this case:
\begin{align}
H(X) = \psi(\sigma((\bm{L_r}X)(\bm{L_c}X)^\top)) \odot \bm{W}   
\end{align}
where $(\bm{L_r}X)(\bm{L_c}X)^{\top} \in \mathbb{R}^{d_r \times d_c}$, $\psi(.)$ is a repeat vector function that repeats its input $\frac{d_{m}}{d_r}$ times on the row axis and $\frac{d_{f}}{d_c}$ times on the column axis. We name this approach the $L^2$ variant, short for Local-Local Gridwise Projection.
\end{minipage}
\begin{minipage}{0.38\linewidth}
    \includegraphics[width=1.0\linewidth]{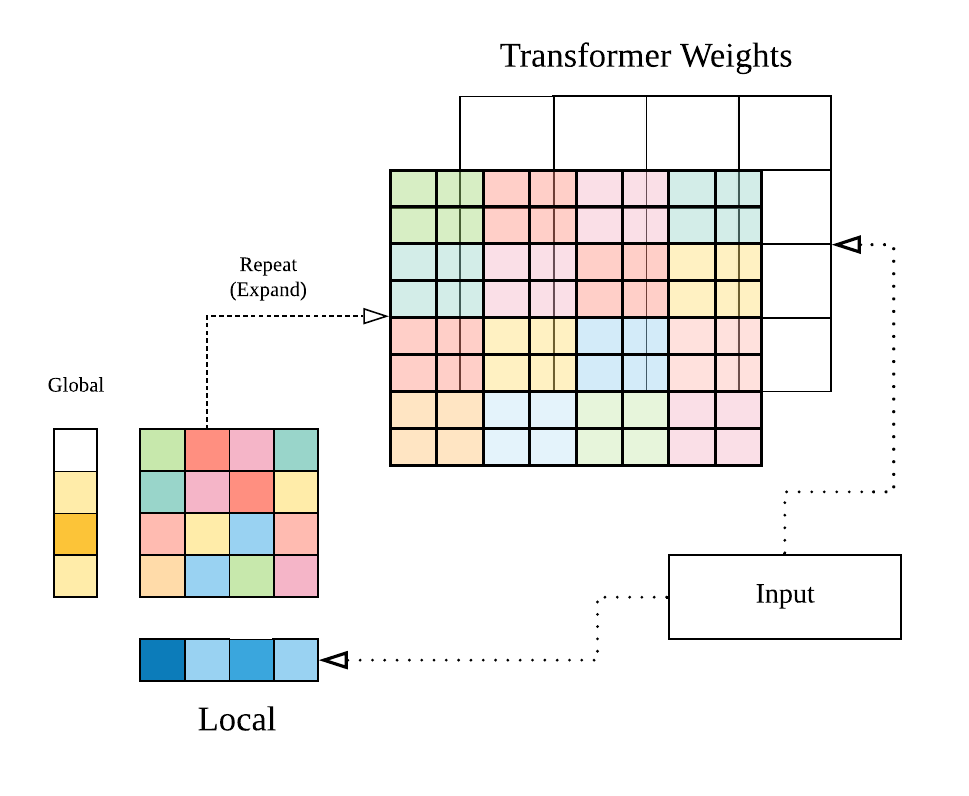}
    \label{fig:detaileddiagram}
    \vspace{-1em}
    \captionof{figure}[]{Detailed Illustration of the proposed Decomposable Gridwise Projections. Two decomposable vectors compose to form a gating matrix which is expanded to construct task-adaptive weight matrices.}
\end{minipage}

\paragraph{Composition between Local and Global Factors} The decomposable grid-wise projections learn $\bm{L_r}$ and $\bm{L_c}$ from $X$, which makes it conditioned on local, instance-wise information. Here, we postulate that it may be beneficial for either $L_r$ or $L_c$ to be a global embedding. By keeping $L_c$ as a global, trainable embedding, this can be formulated as:
\begin{align}
H(X) = \psi(\sigma((\bm{L_r}X)\bm{G_c}^\top)) \odot \bm{W}   
\end{align}
where $\bm{G_c} \in \mathbb{R}^{d_f}$. In this case, $\bm{L_{r}}$ is conditioned from $X$, the specific input sample. On the other hand, $G_c$ remains consistent across all input samples. Hence, the outer product is essentially a rich dyadic composition between local and global factors. 
\paragraph{Local-Global and Global-Local} It is easy to see that there are two ways of composing $L$ and $G$. The above method considers the Local-Global approach where the fan-in uses a local hypernetwork and the global part uses a trainable embedding. An alternative that flips this around to use a Global-Local composition is evaluated in our experiments. Namely, this can be expressed as:
\begin{align}
H(X) = \psi(\sigma((\bm{G_r}(\bm{L_c}X)^\top)) \odot \bm{W}   
\end{align}

\subsection{Multi-Task Fine-tuning of Pretrained Transformers}
Recall that Transformer models \citep{vaswani2017attention} are largely composed of feed-forward transformation layers. We make the following modifications to the Transformer model to equip it with HyperGrid. Note that while our considerations may be designed with T5 \citep{raffel2019exploring} in mind, these findings are expected to transfer to other pretrained models. 
\paragraph{HyperGrid Controlled Feed-forward Layers} We opt to inject HyperGrid at the position-wise feed-forward layers of the Transformer models. More specifically, we equip the second positional FFN after the ReLU activations with HyperGrid. There are several reasons for doing so. In most Transformer implementations, the fan out of this layer is typically scaled up to very large values \citep{raffel2019exploring}. Hence, it is imperative that influence on this layer would benefit the Transformer model the most substantially. Second, early experiments on both of the positional feed-forward layers yielded no substantial improvements. Hence, we opt to only modify the second positional FFN of the Transformer model. Third, in lieu of recent work that downplays the effectiveness of $QKV$ transformations \citep{kitaev2020reformer,tay2020synthesizer}, we do not attempt to apply HyperGrid to the self-attention projections. 

\begin{minipage}{0.5\linewidth}
\begin{figure}[H]
    \centering
    \includegraphics[width=1.0\linewidth]{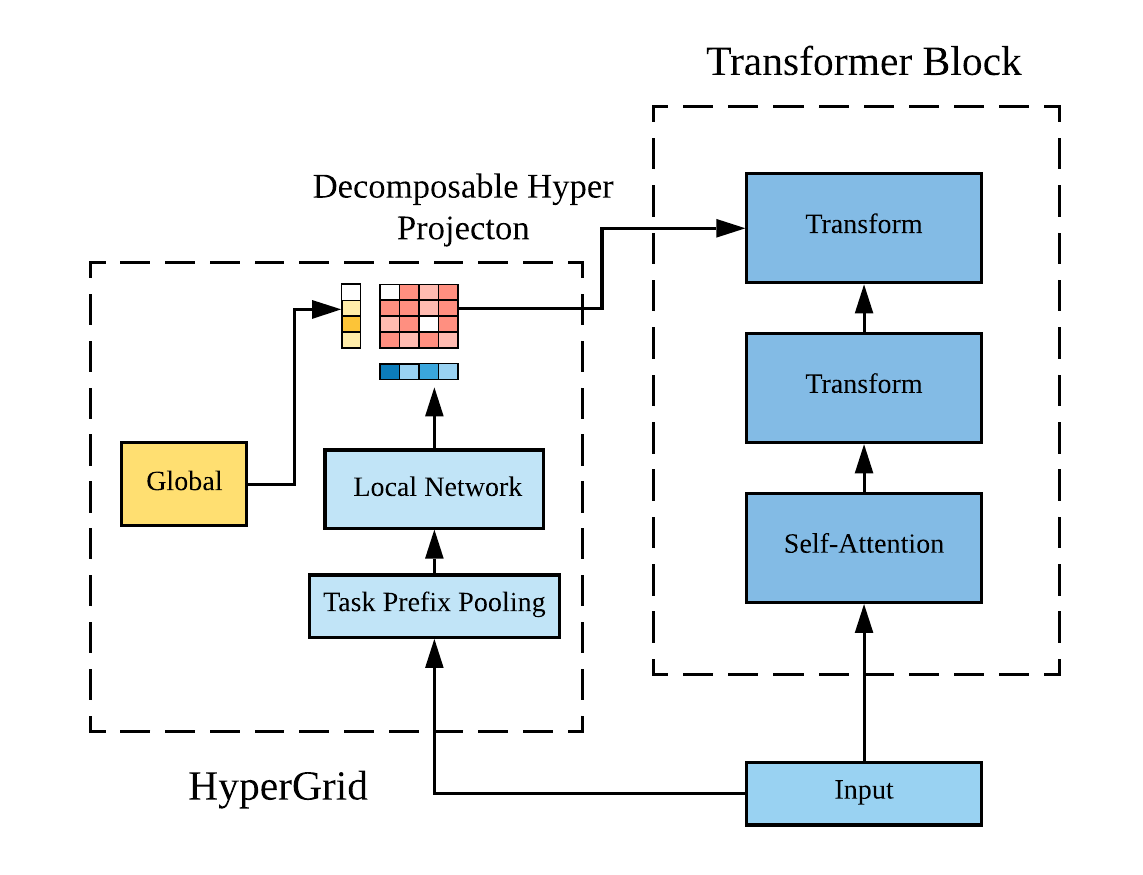}
    \caption{Illustration of the proposed HyperGrid architecture.}
    \label{fig:my_label}
\end{figure}
\end{minipage}
\begin{minipage}{0.5\linewidth}

\paragraph{Task Conditioned HyperGrid for Sequential Inputs} The earlier introduction to the proposed method considers $X$ to be a single feature vector. In practical NLP applications, we are interested in sequential inputs, i.e., $X \in \mathbb{R}^{\ell \times d_m}$. To deal with this, we simply take a pooling $P(.)$ of $X$ that maps $\mathbb{R}^{\ell \times d_m} \rightarrow \mathbb{R}^{d_m}$. For simplicity, we find that a first token pooling works well. Coincidentally, this corresponds to the \textit{prefix} token in the T5 model which provides task information to the model. In our early experiments, we found that an average or sum pooling did reasonably well but did not yield substantial gains over simply using the prefix token. The task prefix token, as the sequence goes through the self-attention layers of the Transformer model, gains context from the neighbouring tokens. Hence, we feel the prefix pooling alone is a reasonable choice.
\end{minipage}
\paragraph{Fine-tuning}
Since our method is primarily developed for multi-task learning, we only use \textsc{HyperGrid} during the fine-tuning stage. This is in similar spirit to Projected Attention Layers (PALS) \citep{stickland2019bert}. We initialize the T5 model using pretrained checkpoints and add additional parameters that are fine-tuned along with the rest of the network. The overall formulation of the HyperGrid-enhanced Transformer can be written as:
\begin{align}
Y_{i} = H_i(X_{i-1},\bm{W_i}) + \bm{W_i}(X_{i-1}) 
\end{align}
where $i$ denotes the layer $i$. We construct a new HyperGrid (with non-shared parameters) for each layer. Since $\bm{W}$ has been pretrained, we also add a residual connection of the original $\bm{W_i}(X_{i-1})$ computation to the mixture.
\paragraph{Parameter Costs} We note that the parameter counts added by \textsc{HyperGrid} are relatively negligible since $d_r$ and $d_c$ are small. In the $LG$ setting, the model adds $d_{m}d_{r} + d_{c}$ parameters at each layer. On the $GL$ setting, the parameter cost added is $d_{r} + d_{f}d_{c}$. The most expensive option is $L^2$ where the added cost is $d_{m}d_{r} + d_{f}d_{c}$. Notably, these costs are often low enough to not appear within the significant digits of large Transformer models.

\section{Experimental Results}

We conduct experiments on GLUE \citep{wang2018glue} and SuperGLUE \citep{wang2019superglue} which are consolidated benchmarks of multiple challenging NLP and NLU tasks. While most of the work in this area has been focused on achieving good task-specific performance, our work focuses on trying to get good performance with a single model on \textbf{all} GLUE and SuperGLUE tasks. Most experiments are conducted on a proportionate mixture of all GLUE and SuperGLUE tasks. This follows the \textit{en\_mix} mixture in the T5 codebase.
\subsection{Datasets and Experimental Setup}
We run most of our experiments using the base T5 setting, which is comprised of $220M$ parameters. We fine-tune for a maximum of $100K$ steps. We initialize our models with the released pretrained checkpoints\footnote{\url{https://github.com/google-research/text-to-text-transfer-transformer}.}. Our implementation is in Mesh Tensorflow \citep{shazeer2018mesh}. We consider the following setups for the baseline T5 model. First, we compare with the T5 results reported in the original\footnote{This model is not directly comparable as they used less pretraining steps. No dev score results on a comparable setup is reported. We report this score for the sake of completeness.} paper \citep{raffel2019exploring}. These results are denoted with T5$^\dagger$. Second, we compare with T5 (PTFT), which stands for pretrain-finetune. In this setup, we fine-tune a T5 model for each task individually following common practice. Finally, we compare with T5 (MTL) which is a fair comparison of T5 without HyperGrid. In this setting, T5 is co-trained and results are reported from a single model checkpoint selected from the best overall GLUE dev score. Note that in the MTL setting, we co-train GLUE and SuperGLUE within the same model. More details can be found in the supplementary material.

\subsection{Experimental Results}
In this section, we discuss the empirical results of our experiments.

\subsubsection{Results on Development Sets} 
Table \ref{gluedev} reports results of our experiments on the GLUE and SuperGLUE benchmark.

\paragraph{Results on GLUE} The first key observation is that the MTL approach is outperformed by PT-FT when using the regular T5 model. This is a well known phenomena and therefore PT-FT is generally adopted when the absolute best score is desired on every single task.  The interesting result is that we are able to come rather close to the performance on PT-FT with our approach. As a result, the T5 (PT-FT) has \textbf{16x} more parameters. To fit both GLUE and SuperGLUE, this would require \textbf{16x} the parameters. Recall that our goal is to bridge the performance of a single model versus multiple models for multiple tasks, we find that this result is considerably successful. Moreover, we observe that our MTL approach outperforms the base T5 using MTL by +0.6\% on average across 8 tasks.  

\paragraph{Results on SuperGLUE} We observe similar trends as on the GLUE benchmark. Naturally, the best model is the PTFT model which involves finetuning a specialized model for each task. The gap between PTFT and MTL is at $74.8$ versus $73.6$. Our approach bridges this gap, improving the MTL score to $74.5$, competitive with the pretrain-finetune methodology. Similar to GLUE, there are also several tasks in which our MTL approach outperforms the PTFT method.

\begin{table}[H]
\scriptsize
    \centering
    \begin{tabular}{l|cccccccccc}
    \hline
Model &	$|\theta|$ &Avg &	CoLA &	SST &	MR & STS & QQP &	MNLI &	QNLI& 	RTE \\
\hline
T5$^\dagger$ &3.2B& 83.4 & 53.8 & 92.7 & 88.9 & 88.0 & 91.6 & 84.4 & 90.5 & 76.3  \\
PTFT & 3.2B & 85.7 &	59.6 &	94.2 &	90.1 &	89.1 &	90.6 &	86.5 &	93.7 &	82.0 \\
\hline
MTL & 0.2B & 85.0 &	57.3&	94.2&	88.6&	89.5&	90.2&	86.2&	93.1&	80.9 \\
Ours ($L^2$) &0.2B & 85.2 &	59.4&	90.6&	90.1&	88.9&	90.3&	86.5&	93.1&	79.1\\
Ours ($LG$) &0.2B & 85.4 &	57.9 &	94.6 &	89.2	&	90.1 	& 	90.3 &		86.7 &	81.2 &	84.2 \\
Ours ($L$) &0.2B&	85.6 &	59.9 &	94.0&	89.1&	89.9 &	90.2	&86.5&	93.1&	81.1 \\
\hline
    \end{tabular}
   \caption{Experimental results on GLUE dev set for base models.}
    \end{table}
    \vspace{-2.5em}
    \begin{table}[H]
    \centering
    \scriptsize
    \begin{tabular}{l|cccccccccc}
       \hline
      Model & 	$|\theta|$ & Avg &	BQ&	CB	&CP&	MultiRC	&Record&	RTE	&WiC&	WSC \\
      \hline
T5$^\dagger$ &  3.2B & 71.4 & 76.6 & 91.2/92.0 & 66.2 &66.1/25.8 &69.1/68.2 & 75.3 & 68.0 &78.6 \\
PTFT & 3.2B & 74.8 & 82.9 &	96.4/92.0	 &	63.0 &	79.1/44.0	&	77.6/76.8 &		83.8 &	71.6 &	73.1	\\
\hline
MTL &0.2B& 73.6 &	81.5&	77.3/83.9	&64.0&	78.2/43.3 &	76.9/76.1&	84.1&	66.9&	74.0 \\
Ours ($L^2)$ & 0.2B & 75.3	 & 82.4&	85.3/91.1&	64.0&	77.8/42.7&	76.8/75.9 &	83.4&	67.1&	80.8 \\
Ours ($LG$) &0.2B & 74.8	& 82.5 &	83.1/89.3	& 64.0	& 77.9/42.8 & 	77.1/76.3	&84.1 &	65.5 & 	78.8 \\ 
Ours ($L$) &0.2B&	74.5&	82.5&	81.5/89.3&	66.0&	78.8/41.0&	76.8/76.0	&85.9&	66.5&	78.8 \\
\hline
    \end{tabular}
     \caption{Experimental results on SuperGLUE dev set for base models. T5$^\dagger$ is reported from \citep{raffel2019exploring} denoted Baseline average. Parameter cost reported is the total parameter cost required to fit GLUE + SuperGLUE. Our multi-task approach bridges the gap between multi-task T5 and pretrain-fine-tuned T5.}
     \label{gluedev}
\end{table}

\subsubsection{Effect of Modeling Choices}
To ascertain the effectiveness of our approach, we test different architectural variants of \textsc{HyperGrid}, along with other architectural variants considered during model development. 

\paragraph{Setup} We evaluate all four model variants of HyperGrid ($L$, $L^2$, $GL$ and $LG$). For the other architectural variants, we were mainly interested to know if a hypernetwork setup (weight gating) is better than gating on the output representations (details to be found in the supplementary material). For the base setting, we ran the baseline T5 model (MTL) four times and reported the mean and standard deviation of the runs. When comparing the performance gain of our method, we compare against the \textbf{max} run of the baseline runs. We report relative performance gains/loss against this max baseline score. We conduct ablation studies on the four composition types on the large models\footnote{Due to the relative increased cost of searching large models, we performed a sparingly low number of ablations on large models.}. 

\begin{minipage}{0.64\linewidth}
    \centering

    \begin{table}[H]
        \scriptsize
    \begin{tabular}{l|ccc}
    \hline
        Model Variant&  GLUE & SuperGLUE & AVG \\
         \hline
         \multicolumn{4}{c}{\textbf{Base Models}} \\
         \hline
         Baseline & 85.03 ($\pm$ 0.087) & 73.77 ($\pm$0.150) & 79.40 ($\pm$0.091) \\
         Baseline (Max) & 85.11 & 73.83 &79.40\\
         \hline
           Local ($L$) & 	85.60 (+0.6\%) &	74.50 (+0.9\%) &	80.05 (+0.8\%)\\
        Local-Local ($L^2$)  &	85.22 (+0.1\%) &	75.30 (+2.0\%) &	80.26 (+1.1\%)\\
        Global-Local ($GL$) & 85.12 (+0.0\%) & 75.00 (+1.6\%)& 80.05 (+0.8\%) \\ 
Local-Global ($LG$) &	85.43 (+0.4\%)  &	74.78 (+1.3\%) &80.10 (+0.9\%) \\
         \hline
         OutGate (Full) & 85.13 (+0.0\%) & 73.31 (-0.7\%) & 79.22 (-0.2\%) \\
         OutGate ($16$) & 84.94 (-0.2\%) & 73.10 (-1.0\%)  & 79.01 (-0.5\%) \\
         OutGate ($32$) & 84.84 (-0.3\%) & 72.93 (-1.2\%)& 78.89 (-0.6\%)  \\
         OutGate ($64$) & 85.07 (-0.0\%) & 74.11 (+0.4\%) & 79.59 (+0.2\%) \\
         \hline
         \multicolumn{4}{c}{\textbf{Large Models}} \\
         \hline
         Baseline & 88.22 & 80.04 & 84.13 \\
         Local ($L$) & 88.07 (-0.2\%) & 80.51 (+0.6\%) & 84.29 (+0.2\%) \\
        Local-Local ($L^2$) & 88.05 (-0.2\%) & 80.68 (+0.8\%) & 84.36 (+0.3\%) \\
         Global-Local ($GL$) & \textbf{88.33} (+0.1\%) & 80.30 (+0.3\%) & 84.32 (+0.2\%) \\
        Local-Global ($LG$) & 88.31 (+0.1\%) & \textbf{81.56} (+1.9\%) & \textbf{84.94} (+1.0\%) \\ 
        \hline
    \end{tabular}
    \caption{Ablation Study} 
    \label{tab:ablation}
    \end{table}
\end{minipage}
\begin{minipage}{0.37\linewidth}
\paragraph{Findings of HyperGrid Variants} 
Table \ref{tab:ablation} reports our key ablation results.  Pertaining to results of the base models, our overall finding is that HyperGrid generally improves performance over the max baseline. Gains are mainly on SuperGLUE while maintaining good performance on GLUE. The overall average gain is about $+1\%$. Amongst the different variants of HyperGrid, the best performing model on this setup is the $L^2$ setup. On the large setting, we find that the $LG$ model performs the best while the $L$ and $L^2$ variants perform similar to the baseline. 
\end{minipage}

\paragraph{Is Output Gating Better?} The other architectural variants (OutGate) do not perform well and generally perform with a net loss in performance as compared to the baseline. As such, we ascertain that gating on weights is more effective than gating on the output representations. This verifies that our hypernetwork-based approach is indeed effective as opposed to simple task-conditioned output gating.

\subsubsection{Performance Gains across Model Sizes}
We investigate the gains of the proposed HyperGrid over the base model on various sizes of the T5 model. For models larger than Base, we train with 64 TPU V3 chips for $200K$ steps and select the best checkpoint for all tasks based on the GLUE score.

\begin{minipage}{0.6\linewidth}
\begin{table}[H]
    \centering
    \scriptsize
    \begin{tabular}{l|lll}
    \hline
    Model / Size  & 	GLUE& 	SuperGLUE& 	AVG\\
    \hline
T5 Base	&  84.99& 	73.55	& 79.27 \\
Ours Base & 85.22 (+0.27\%) &	75.30 (+2.7\%) &	80.26 (+1.3\%) \\ 
\hline
T5 Large& 	88.22 & 	80.04& 	84.13 \\
Ours Large& 	88.31 (+0.1\%) 	& 81.56 (+1.9\%) & 	84.94 (+1.0\%)\\
\hline
T5 3B	& 89.53& 	84.22& 	86.87 \\
Ours 3B & 	\textbf{89.67} (+0.2\%)& 	\textbf{85.75}  (+1.8\%) & 	\textbf{87.71} (+1.0\%) \\
\hline
    \end{tabular}
    \caption{Effect of HyperGrid on Multi-Task T5 on all model sizes. HyperGrid improves multi-task co-training consistently overly different model sizes. Improvement over SuperGLUE is greater than GLUE.}
    \label{tab:size}
\end{table}
\end{minipage}
\begin{minipage}{0.38\linewidth}
\paragraph{Findings} Table \ref{tab:size} reports results of GLUE and SuperGLUE scores (and their macro-average). We find that performance gains on SuperGLUE averages is reasonably good ($+1.9\%$ on Large). The model still outperforms the vanilla model on GLUE with marginal performance gains. Overall, on a macro-average of $18$ tasks, we find an overall $+1.0\%$ improvement across three sizes. These results show that performance gains scale with model size. 
\end{minipage}

\subsubsection{Effect of Grid Size on Performance}
We investigate the effect of Grid size (fan-in and fan-out) of our proposed HyperGrid method. The purpose of this experiment is to discover how fine-grained or coarse-grained the hypernetwork should be. Notably, smaller values of $d_r,d_c$ signify a more coarse-grained control of the Transformer weights. 
\paragraph{Setup} We searched $d_r$ (fan-in) and $d_c$ (fan-out) in the ranges of $\{4, 8,16,32,128,256\}$ and $\{8,16,32,128,256\}$ respectively and report the results on GLUE + SuperGLUE (macro-average) by varying a single value. When varying $d_r$, we took the average of all $d_c$ runs and plot the max, mean and min. Likewise, when varying $d_c$, we took the average of all $d_r$ runs and plot max, mean and average. We report scores across the $L^2$, $LG$, and $GL$ variants of HyperGrid.

\begin{figure}[]
\small
\begin{minipage}{0.33\linewidth}
    \centering
    \includegraphics[width=0.98\linewidth]{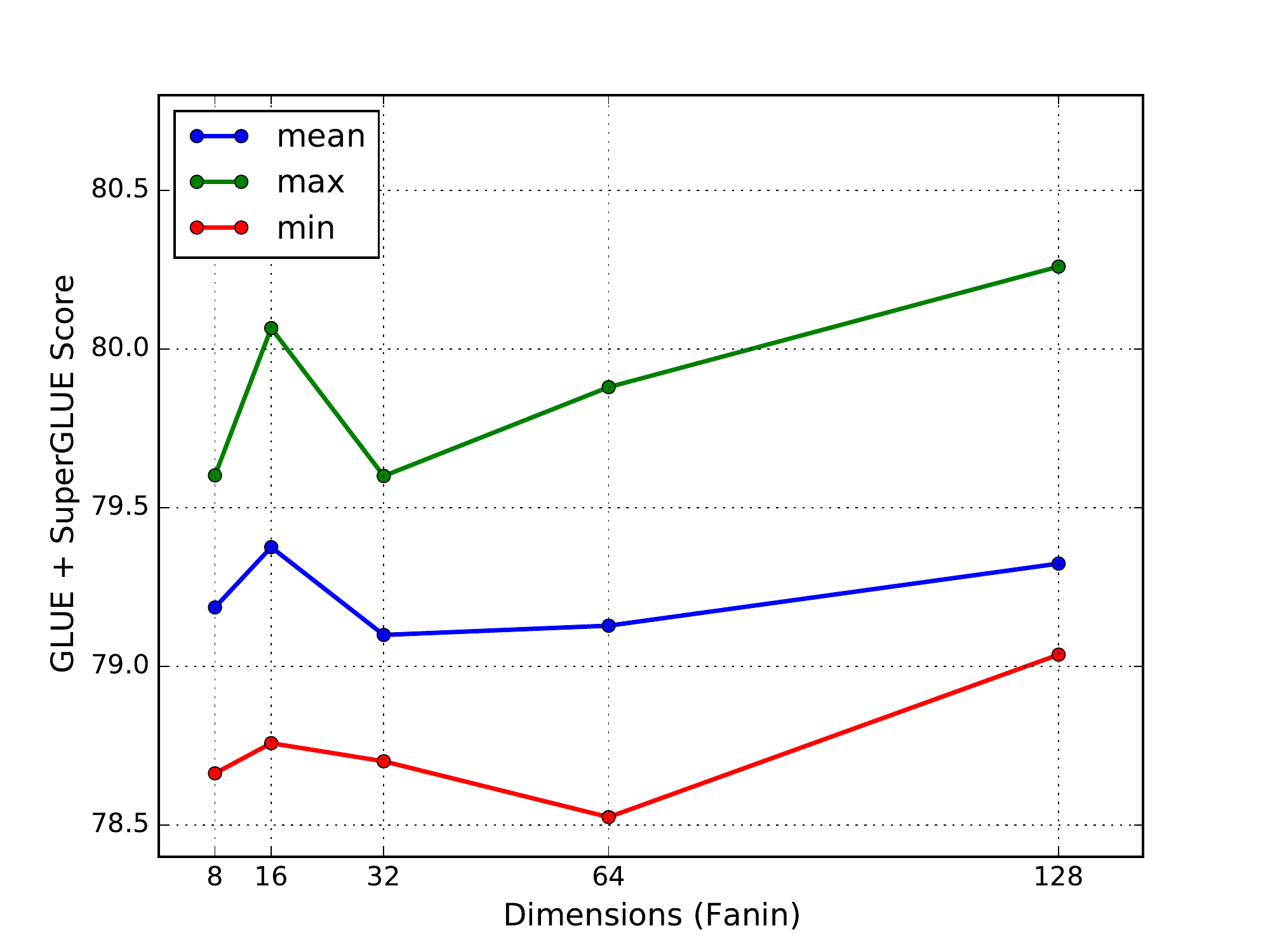}
    \caption{fan-in on $L^2$ setting.}
    \label{fig:my_label}
    \end{minipage}
\begin{minipage}{0.33\linewidth}
    \centering
    \includegraphics[width=0.98\linewidth]{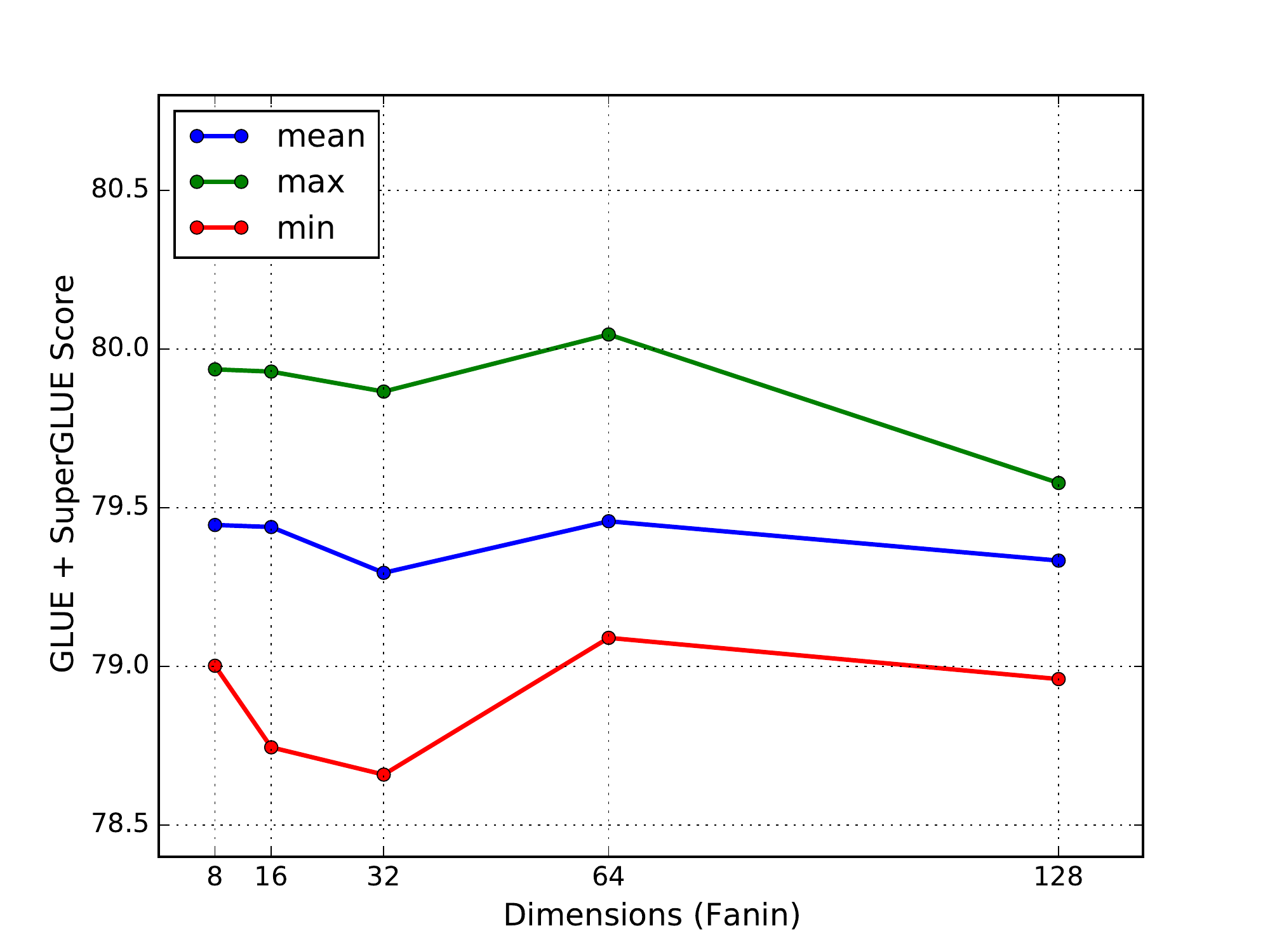}
  \caption{fan-in on $LG$ setting.}
    \label{fig:my_label}
    \end{minipage} 
\begin{minipage}{0.32\linewidth}
    \centering
    \includegraphics[width=0.98\linewidth]{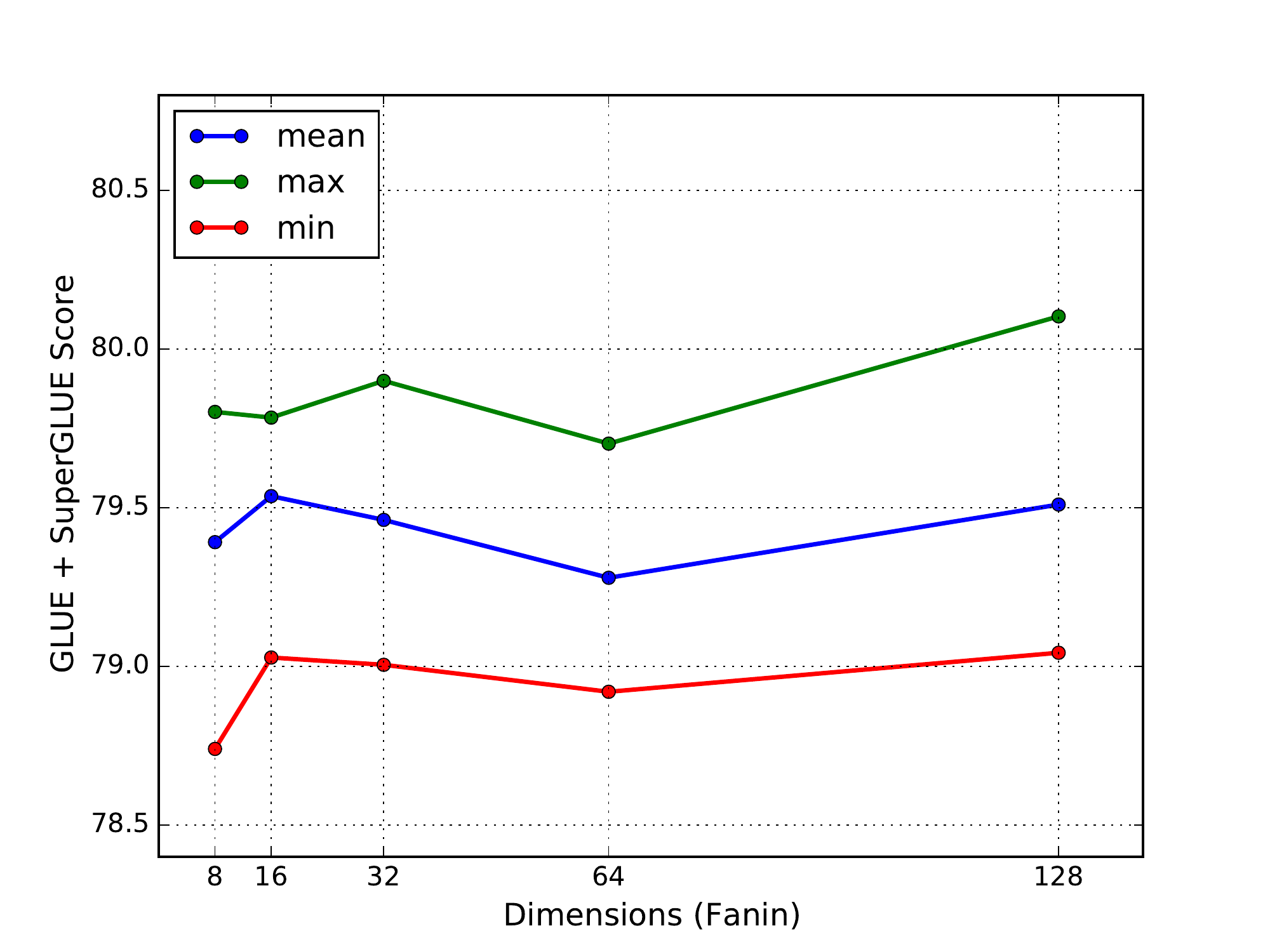}
    \caption{fan-in on $GL$ setting.}
    \label{fig:my_label}
    \end{minipage}
\begin{minipage}{0.33\linewidth}
    \centering
    \includegraphics[width=0.98\linewidth]{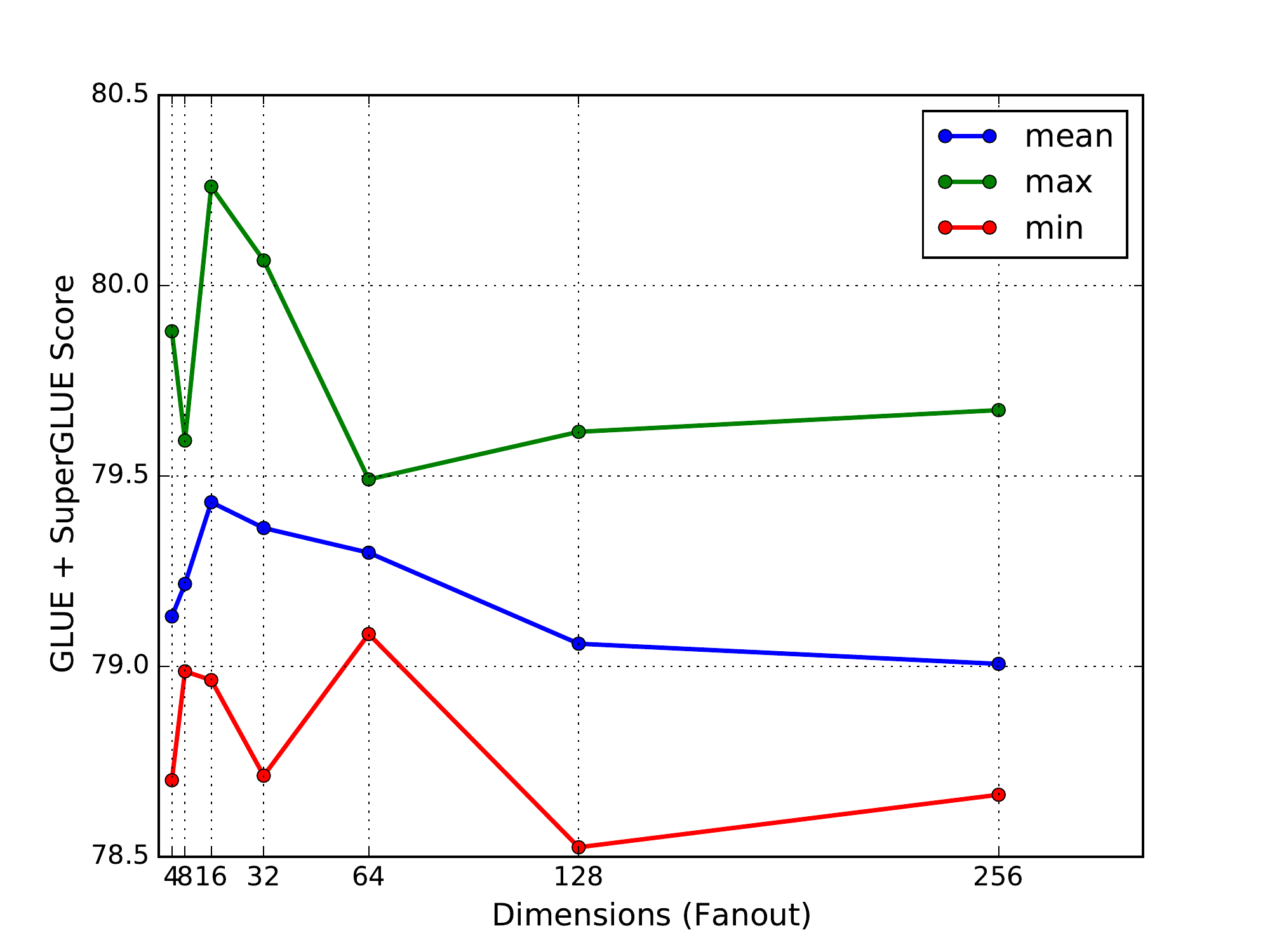}
  \caption{fan-out on $L^2$ setting.}
    \label{fig:my_label}
    \end{minipage}
\begin{minipage}{0.33\linewidth}
    \centering
    \includegraphics[width=0.98\linewidth]{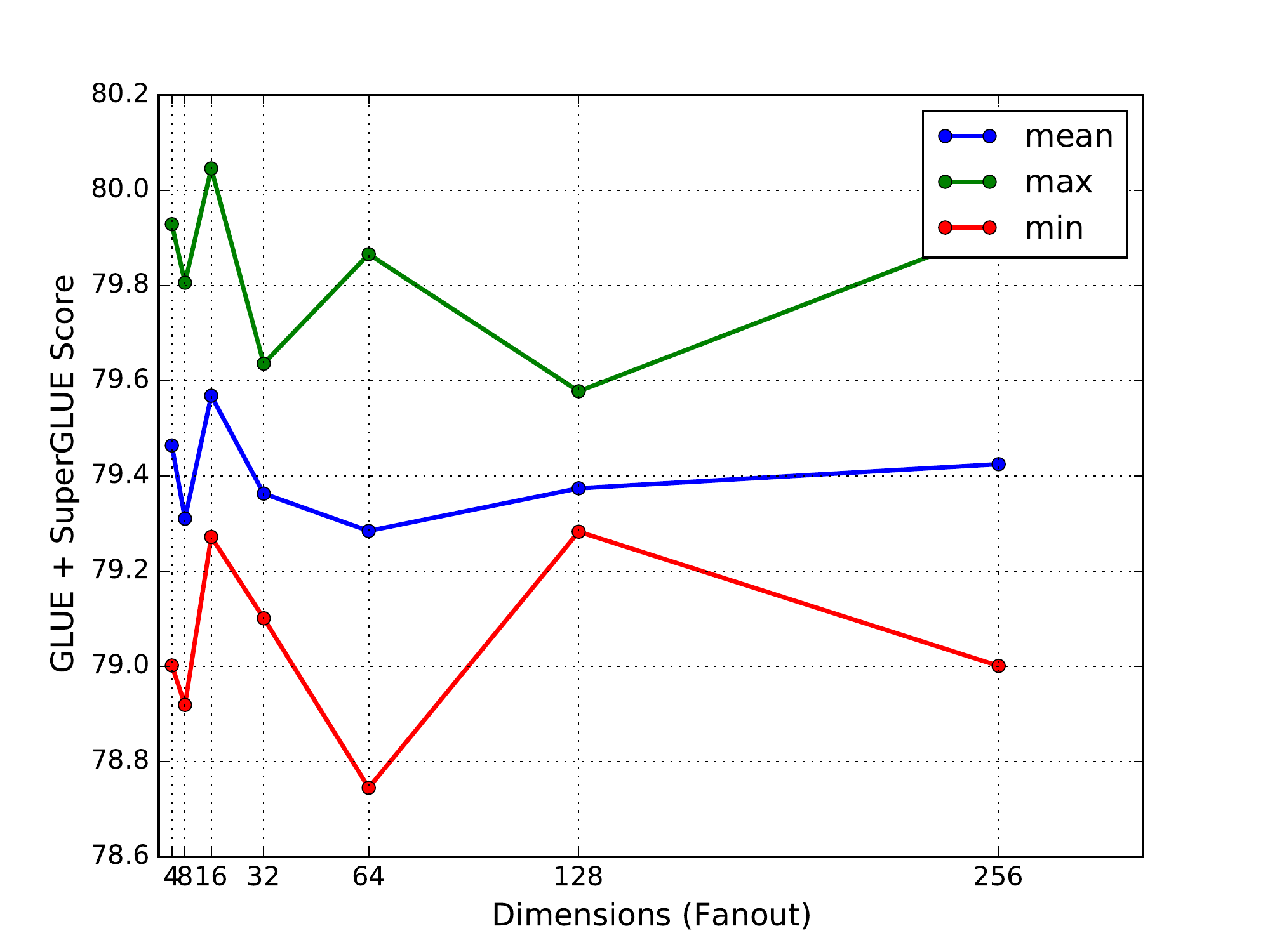}
  \caption{fan-out on $LG$ setting.}
    \label{fig:my_label}
    \end{minipage} 
\begin{minipage}{0.33\linewidth}
    \centering
    \includegraphics[width=0.98\linewidth]{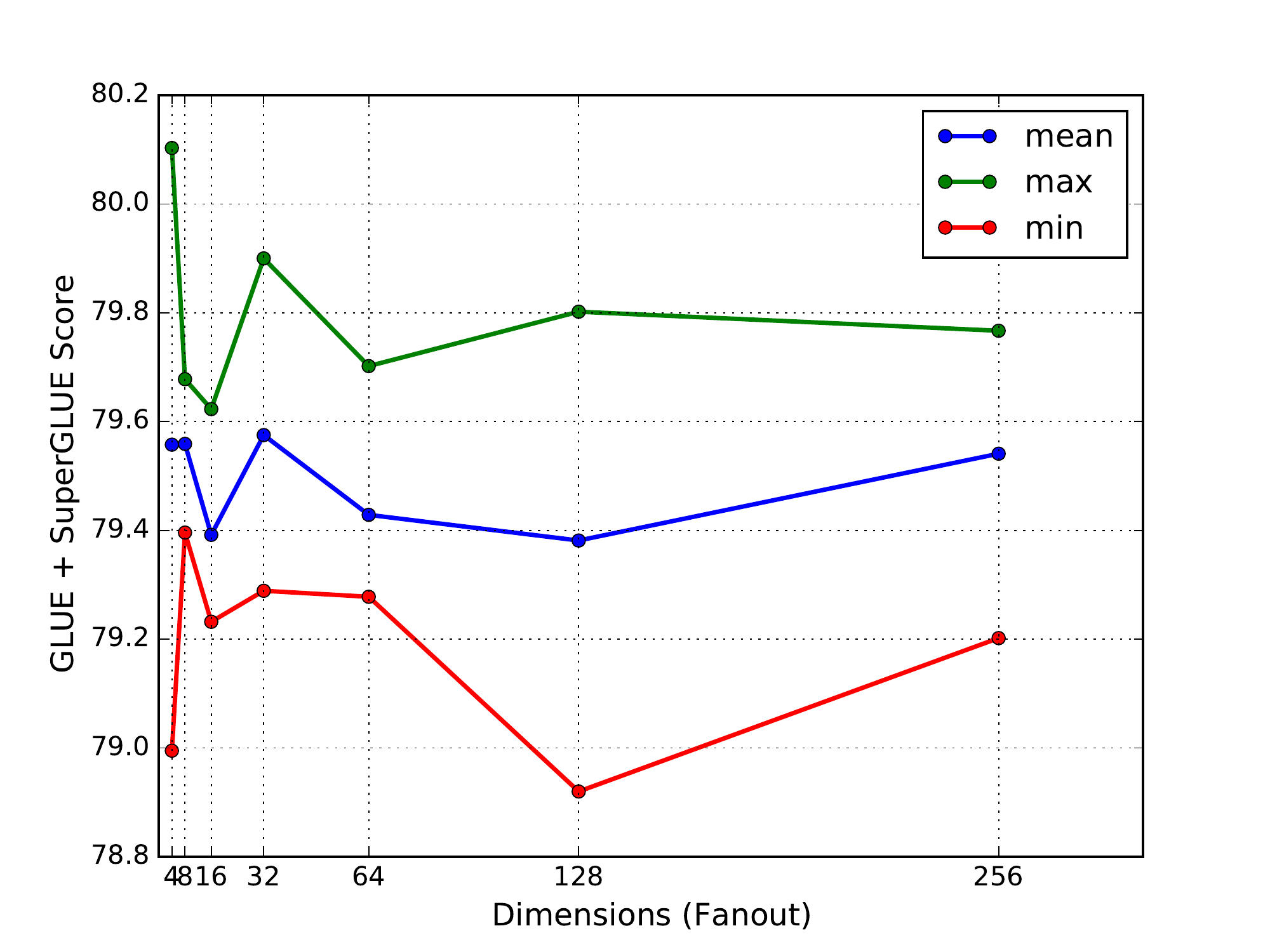}
  \caption{fan-out on $GL$ setting.}
    \label{fig:my_label}
    \end{minipage}    
\caption{Effect of Grid size (fan-in and fan-out) on performance on GLUE and SuperGLUE.}
\label{fig:grid}
\end{figure}
\paragraph{Findings pertaining to Grid Size} Figure \ref{fig:grid} illustrates performance across varied grid sizes. From the charts, we observe that a clear trend exists. For most settings, a small fan-out ($d_c$) works well (e.g., $32$) as noted by many spikes around this region. For fan-in ($d_r)$ a smaller value also works well. However, performance gets better at higher fan-out $d_c$ values again (e.g., $>128$). Trends are quite consistent across all three variations that we considered. These results suggest that a more coarse grid may be more effective, as the regions within the grid become larger.

\begin{table}[H]
\scriptsize
    \centering
    \begin{tabular}{l|ccccccccccc}
    \hline
    \hline
Model &	$|\theta|$ &Avg &	CoLA &	SST &	MR & STS & QQP &	MNLI &	QNLI& 	RTE & WNLI\\
\hline
	\hline
BERT$^*$	& -	&	80.5	&	60.5	&	94.9	&	84.5	&	86.5	&	89.3	&	86.7	&	92.7	&	70.1	&	65.1 \\
RoBERTa$^*$		& -	&	88.1	&	67.8	&	96.7	&	89.8	&	91.9	&	90.2	&	90.8	&	95.4	&	88.2	&	89.0 \\
ALBERT$^*$		&	-	&- 	&	69.1	&	97.1	&	91.2	&	92.0	&	90.5	&	91.3	&	-	&	89.2	&	89.0 \\
XLNet$^*$		& -	&	-	&	70.2	&	97.1	&	90.5	&	92.6	&	90.4	&	90.9	&	-	&	88.5	&	89.1\\
ELECTRA$^*$		& 5B	&	89.4	&	71.7	&	97.1	&	90.7	&	92.5	&	90.8	&	91.3	&	95.8	&	88.5	&	92.5	\\
T5 	(3B) 	&	48B	&	88.5	&	67.1	&	97.4	&	90.0	&	89.8	&	82.1	&	91.3	&	96.3	&	91.1	&	89.7\\ 
T5 (11B)	&	176B		&89.7	&	70.8	&	97.1	&	90.0	&	92.1	&	82.5	&	90.9	&	96.7	&	92.5	&	93.2\\
\hline
Ours (3B)	&	3B	&	88.2	&	65.6	&	97.5	&	89.0	&	91.6	&	81.9		&90.9	&	95.9	&	90.1	&	89.7 \\
Ours (11B) & 11B		&	89.4	&	69.0		&97.6		&89.2		&92.6	&	82.0	&	91.3	&	96.4	&	91.5	&	93.2 \\
\hline
\hline
    \end{tabular}
    \caption{Test set performance on GLUE \citep{wang2018glue}. Models with $*$ are large ensembles. All models are single-tasked fine-tuned except ours. Parameter costs are reported considering ensembles and cost required to fit all of GLUE and SuperGLUE.}
\end{table}
\vspace{-2.5em}
\begin{table}[H]
    \centering
    \scriptsize
  \begin{tabular}{l|cccccccccc}
  \hline
  \hline
        Model & 	$|\theta|$ & Avg &	BQ&	CB	&CP&	MultiRC	&Record&	RTE	&WiC&	WSC \\
\hline
BERT++ &	2.7B&	71.5&	79.0&	84.8/90.4&	73.8 & 70.0/24.1& 	72.0/71.3	& 79.0&	69.6	& 64.4 \\
RoBERTa&	56B&	84.6&	87.1 & 90.5/95.2&	90.6 &84.5/52.5 &	90.6/90.0 &	88.2&	69.9&	89.0  \\
T5 (3B) &	48B &	86.4&	89.9&	90.3/94.4&	92.0	&86.8/58.3	& 91.2/90.4	& 90.7 &72.1&	90.4 \\
T5 (11B) &	176B	&88.9&	91.0 &	93.0/96.4&	94.8&	88.2/62.3&	93.3/92.5&	92.5&	76.1&	93.8\\
\hline
Ours (3B)&	3B&	84.7&	89.2&	81.7/90.4	& 89.6&	86.6/58.7&	91.1/90.3&	90.8&	70.6&	87.7 \\
Ours (11B)&	11B	& 87.7&	90.7&	85.5/92.0 &	94.0 & 87.9/61.7&	93.3/92.6 &	91.5&	74.6 &	92.1\\
\hline
\hline
    \end{tabular}
    \caption{Test set performance on SuperGLUE \citep{wang2019superglue}. Our MTL approach achieves competitive performance to the state-of-the-art with a single multi-task model. Parameter costs refers to total number of parameters used to fit all GLUE and SuperGLUE tasks}
    \label{tab:my_label}
\end{table}

\subsubsection{Performance on Test Set}
For our final runs, we submit our model predictions to the GLUE and SuperGLUE test servers. 
\paragraph{Setup} We fine-tune a 3B and 11B model in multi-task\footnote{Since we did not co-train with the WNLI dataset due to issues stated in \citep{raffel2019exploring}, we simply report T5 results on WNLI. To be fair, we ignore WNLI parameter counts for all baseline models. } setup (GLUE + SuperGLUE) using T5 pre-trained checkpoints. Since this is a relatively expensive run,  we only train the MTL \textsc{HyperGrid} model once using a $32 \times 128$ grid with the $LG$ (local-global) setting. To avoid an excessive number of submissions to the test server, we do not evaluate our MTL baselines since it has been shown from dev scores that our MTL approach outperforms the MTL T5. For GLUE, we compare against baselines reported in \citep{clark2020electra} which includes models such as BERT \citep{devlin2018bert}, ALBERT \cite{lan2019albert}, RoBERTa \citep{liu2019roberta} and XLNet \citep{yang2019xlnet}. Note that all these models are not only ensembles but heavily rely on task specific fine-tunining strategies. More details can be found in the supplementary material.

\paragraph{Results on Test Set} We find that our MTL approach can achieve highly competitive results on both GLUE and SuperGLUE. Our model achieves a strong performance of $87.7$ on SuperGLUE, just $1.2\%$ shy of the state-of-the-art while having $16$ times less \textit{total} parameters. On GLUE, the performance gap is even smaller, almost matching the T5 model at $89.4$ versus $89.7$. The gap on the base model remains similar at $88.2$ versus $88.5$. On SuperGLUE, our 3B model achieves $84.7$, a respectable score that matches the performance of RoBERTa ensembles fine-tuned individually with task specific tricks \citep{liu2019roberta}.

\section{Conclusion}
We proposed Grid-wise Decomposable Hyper Projections (\textsc{HyperGrid}), a hypernetwork-based projection layer for efficient fine-tuning of multi-task Transformers. We learn and fit all GLUE and SuperGLUE tasks within the same set of model parameters and achieve competitive results to the same state-of-the-art model that is specially and individually fine-tuned on each and every tasks. On GLUE/SuperGLUE, this efficient multi-tasking method results in $16$ times parameter savings.  

\newpage
\section{Broader Impact}
This paper proposes a task-conditional method for fine-tuning of large generative Transformer models. 

\paragraph{Impact on Multi-Task Learning} While we apply this on natural language understanding tasks, this can, in principle, be applied to any group of supervised machine learning tasks in a multi-task setting. Ultimately, the goal is to reduce the number of served models in a production environment by training as many tasks as possible within a single model. This has the potential for reducing energy consumption, as we no longer need to expend computational resources to fine-tune and serve different models for every possible task. 
\paragraph{Impact on Transformer Research} 
This work also impacts Transformer architecture research as the extended fine-tuned architecture can be considered a Transformer variant. This paper shows the promise of architectural improvements for task-conditional feed-forward layers. This may spur future research on learning task-conditional Transformer models.
\paragraph{Impact on Natural Language Understanding} This paper shows that multiple natural language understanding tasks can be fit using a single model while achieving highly competitive results. It also addresses the issue where task-specific fine-tuning tricks and aggressive ensemble learning may be infeasible in practice.

\bibliographystyle{plainnat}
\bibliography{references}

\newpage

\section{Supplementary Material}

\subsection{Datasets}

\subsubsection{GLUE} The datasets in GLUE are CoLA (Corpus of Linguistic Acceptability) \citep{warstadt2018neural}, Sentiment Treebank SST-2 \cite{socher2013recursive}, Microsoft Research Paraphrase Corpus (MRPC) \citep{dolan2005automatically}, QQP (Quora Question Pairs) \citep{WinNT}, Semantic Textual Similarity Benchmark (STSB) \citep{cer2017semeval}, MNLI (Multi-Genre Natural Language Inference) \cite{N18-1101}, QNLI \citep{rajpurkar2016squad}, RTE \citep{dagan2005pascal}, Winograd Schema Challenge WNLI \citep{levesque2012winograd}. More details can be found at \url{https://github.com/tensorflow/datasets/blob/master/docs/catalog/glue.md}.

\subsubsection{SuperGLUE}
The datasets in SuperGLUE \citep{wang2019superglue} are BoolQ (Boolean Questions) \citep{clark2019boolq}, CB (Commitment Bank) \citep{demarneff_simons_tonhauser_2019}, CoPA \citep{roemmele2011choice} (Choice of Plausible Alternatives), MultiRC (Multi-Sentence Reading Comprehension Dataset) \citep{MultiRC2018}, Record (Reading Comprehension with Commonsense Reasoning) \citep{zhang2018record}, RTE (Recognizing Textual Entailment) \citep{dagan2005pascal,bar2006second,giampiccolo2007third,bentivogli2009fifth}, Word-in-Context (WiC) \citep{DBLP:journals/corr/abs-1808-09121}, and WSC (Winograd Schema Challenge) \citep{levesque2012winograd}. We use Tensorflow datasets for loading and preprocessing these datasets. More details can be found at \url{https://github.com/tensorflow/datasets/blob/master/docs/catalog/super_glue.md}. 

\subsection{Experiment Settings}
This section describes most of the hyperparameter settings for our experiments. 
\paragraph{Experiments for Base Models} For all experiments with base models, we train models for $100K$ steps with a batch size of $128$. We use the \textit{en\_mix} mixture which samples each task proportionately to the number of examples in the dataset. Learning rate is a constant $0.001$ with Adafactor \citep{shazeer2018adafactor}. All results for baselines are reported with scores at the last checkpoint. During fine-tuning, the embeddings are not fine-tuned. Experiments are run with 16 TPU V3 chips and are typically completed in about $8$ to $10$ hours.

\paragraph{Experiments with Large Models} We increased the search for large models to $200K$ steps pick the best checkpoint for all models based on the best GLUE score. Experiment and hyperparameter settings remain identical although we use $64$ TPU V3 chips for finetuning which typically take about $12$ hours to complete. 

\paragraph{Experiments with 3B and 11B Models} For the large models, we only use $1-2$ HyperGrid configurations 32x128 or 32x256 in $LG$ mode for finetuning the model. We submit each model only once to the leaderboard\footnote{Discounting submissions that turn out to be incomplete or error submissions.}. Finetuning hyperparameters remain identical. We pick a single checkpoint based on the best GLUE score. Finetuning for the $3B$ model is using $64$ TPU V3 chips and the $11B$ model is fine-tuned with $128$ TPU V3 chips. 

\subsection{Comparing with Output Gating}
One of the model architecture variants we compared with is Output Gating. It can be formulated as:
\begin{align}
Y = \max(\bm{W}x +b, 0) \odot (\sigma(\bm{U}X)\bm{1}^{\top})
\end{align}

Comparing to the HyperGrid, which gates the weights in the Relu layer, output gating directly gates the Relu layer outputs. We can  apply either the basic projection method (Equation (2)), or the grid-wise projection method with block-wise projection on layer outputs. 

There are two key differences: (1) Output Gating applies sigmoid gating on Relu layer outputs, while HyperGrid applies sigmoid gating on weights before the Relu function. Output gating is similar to the Mixture-of-Expert architecture while concatenating the expert outputs. (2) Based on this formulation, the full grid-based projection cannot be applied to output gating.

\end{document}